\documentclass[letterpaper, 10 pt, conference]{ieeeconf}
\usepackage{cite}
\usepackage{amsmath}
\usepackage{graphicx}
\usepackage{tabularx}
\usepackage{array}
\usepackage{amssymb}
\usepackage{cuted}
\usepackage{caption}
\usepackage{hyperref}

\usepackage{booktabs}
\usepackage{arydshln}
\usepackage{makecell}
\usepackage{multirow}
\usepackage{microtype} 
\usepackage{flushend}

\IEEEoverridecommandlockouts                           
\title{\LARGE \bf
From Watch to Imagine: Steering Long-horizon Manipulation via Human Demonstration and Future Envisionment
}

\author{\href{https://yipko.com/about/}{Ke Ye}$^{1*}$, \href{https://jiaming-zhou.github.io/}{Jiaming Zhou}$^{1*}$, \href{https://github.com/AaronQyf}{Yuanfeng Qiu}$^{1*}$, \href{https://www.jiayi-liu.cn/}{Jiayi Liu}$^{1}$, \href{https://github.com/zhoush210}{Shihui Zhou}$^{1}$, \href{https://kunyulin.github.io/}{Kun-Yu Lin}$^{2}$ and \href{https://junweiliang.me/}{Junwei Liang}$^{1,3\dagger}$%
\thanks{$*$ Indicates co-first authors.}%
\thanks{$^{1}$K. Ye, J. Zhou, Y. Qiu, J. Liu, S. Zhou and J. Liang are with The Hong Kong University of Science and Technology (Guangzhou), Guangzhou, China}%
\thanks{$^{2}$K. Lin is with The University of Hong Kong, Hong Kong SAR, China}%
\thanks{$^{3}$J. Liang is also with The Hong Kong University of Science and Technology, Clear Water Bay, Hong Kong SAR, China}%
\thanks{ Project Lead: Jiaming Zhou. $\dagger$ Corresponding author: Junwei Liang. {\tt\small junweiliang@hkust-gz.edu.cn}}%
}

\begin{document}

\maketitle
\thispagestyle{empty}
\pagestyle{empty}
\begin{strip}
\vspace*{-25mm}
\centering
\includegraphics[width=1.0\textwidth]{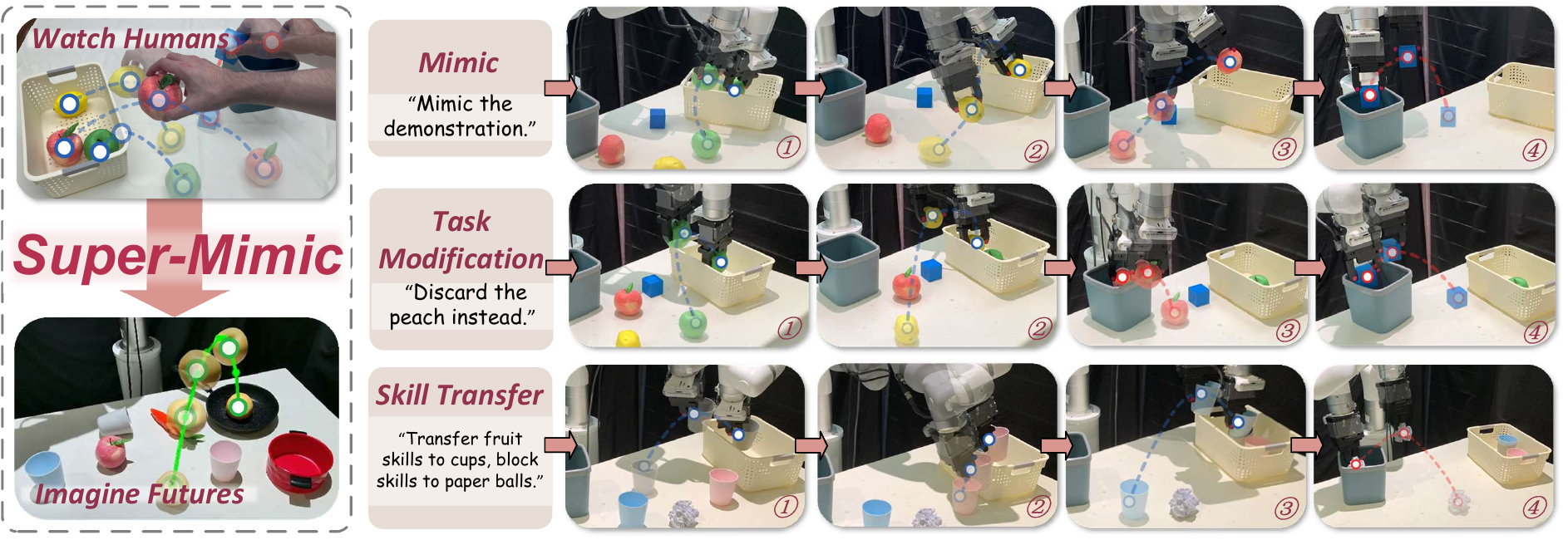}
\captionsetup{type=figure}
\caption{We proposed \textbf{Super-Mimic}, a system that goes beyond simple imitation of human actions, enabling task modifications and new skill transfers, and operates in a zero-shot manner without requiring task-specific training data.}
\label{fig:head}
\end{strip}
\begin{abstract}
Generalizing to long-horizon manipulation tasks in a zero-shot setting remains a central challenge in robotics. Current multimodal foundation based approaches, despite their capabilities, typically fail to decompose high-level commands into executable action sequences from static visual input alone.
To address this challenge, we introduce \textbf{Super-Mimic}, a hierarchical framework that enables zero-shot robotic imitation by directly inferring procedural intent from unscripted human demonstration videos.
Our framework is composed of two sequential modules. First, a Human Intent Translator (HIT) parses the input video using multimodal reasoning to produce a sequence of language-grounded subtasks. These subtasks then condition a Future Dynamics Predictor (FDP), which employs a generative model that synthesizes a physically plausible video rollout for each step. The resulting visual trajectories are dynamics-aware, explicitly modeling crucial object interactions and contact points to guide the low-level controller.
We validate this approach through extensive experiments on a suite of long-horizon manipulation tasks, where Super-Mimic significantly outperforms state-of-the-art zero-shot methods by over 20\%. These results establish that coupling video-driven intent parsing with prospective dynamics modeling is a highly effective strategy for developing general-purpose robotic systems. More details and videos can be found at: \href{https://yipko.com/super-mimic}{yipko.com/super-mimic}

\end{abstract}

\section{INTRODUCTION}

The pursuit of generalized robotic manipulation aims to develop systems capable of performing a wide array of tasks in diverse, unstructured environments without task-specific training. These systems must comprehend high-level instructions, perceive visual scenes, and generate feasible action plans in a zero-shot manner.
The recent success of multimodal large language models (MLLMs)\cite{Qwen2.5-VL, hurst2024gpt, seed2025seed1_5vl}, pre-trained on vast internet-scale datasets, has demonstrated their inherent capabilities in perception and reasoning\cite{comanici2025gemini,duan2024aha, wang2024vlm}. 
Leveraging these models as core components presents a promising pathway toward building such general-purpose robotic systems.

However, existing zero-shot approaches \cite{huang2023voxposer, huang2024rekep, liu2024moka} based on foundation models are often limited to simple, short-horizon tasks, such as \textit{picking up a cup}. 
While some works \cite{ahn2022can, wu2023tidybot} have attempted more complex, long-horizon tasks like \textit{tidying up the floor}, their planning and execution rely heavily on textual descriptions. 
This is problematic because it forces MLLMs to infer complex action sequences from simple descriptions and cluttered visual scenes, a process that is often unreliable and difficult to control. 
Furthermore, providing detailed instructions for complex, long-horizon tasks is laborious, and the constituent steps of a long sequence often cannot be explicitly segmented or described with language.

We believe that for a robot to achieve human-level proficiency in complex tasks, it should learn from a more direct and richer source of instruction: human demonstrations. 
To this end, we propose that unscripted human videos serve as a more effective and natural high-level instruction modality. 
The rich visual information contained within these videos implicitly encodes the necessary subtasks and corresponding manipulation skills required to complete long-horizon tasks. 
Therefore, we introduce the \textit{Human Intent Translator (HIT)}. This module first efficiently extracts key action frames from the human video.
It then leverages MLLMs to build a unified intent translation process that generates a transferable subtask plan based on the parsed keyframe sequence and the robot's current visual observation. 
This process can also be refined with additional text prompts for targeted plan modification, enabling effective imitation of the long-horizon task demonstrated in the human video.

Furthermore, in a zero-shot setting, the successful execution of each subtask is critical for completing a long-horizon task. 
Prior works often struggle to extract robust dynamic semantics for subtask execution using only MLLMs, which primarily excel at static scene understanding. 
For instance, MOKA\cite{liu2024moka} proposes 2D keypoints on a static image to form a trajectory. 
Similarly, other approaches \cite{huang2023voxposer, huang2024rekep, liang2022code, zeng2022socratic, singh2022progprompt, huang2022inner} attempt to regress end-effector poses directly from static observations, often failing to capture the nuances of object interactions. 
In this work, to ensure robust subtask planning in open-world scenarios, we introduce the \textit{Future Dynamics Predictor (FDP)}, which employs video generation models\cite{nvidia2025cosmos, zhen2025tesseract, wan2025, kong2024hunyuanvideo, polyak2024movie} to prospectively \textit{imagine} the execution of each subtask. 
This practice not only preserves the spatial structure and fine details from the robot's current visual observation but also accurately infers the complex contact dynamics and constraints between multiple objects. 
These imagined futures serve as a dense, physically-grounded reference for the subsequent control phase. 
For low-level control, we extract depth estimates and key trajectories from these generated videos, which are then translated into executable 3D target poses and control signals. 
This provides fine-grained manipulation and stable performance, particularly in scenes involving multi-object interactions.

In this paper, we present \textbf{Super-Mimic}, a framework that combines the HIT module for task planning and the FDP module for subtask execution to tackle zero-shot long-horizon robotic manipulation (Fig.~\ref{fig:head}). 
Our key contributions are:

\noindent 1) We propose a novel hierarchical framework to solve zero-shot long-horizon manipulation, by using human demonstration videos to guide high-level task planning and video generation models to inform robust low-level control.

\noindent 2) We introduce the HIT module, which parses complex human actions into transferable, language-guided robotic plans, and the FDP module, which imagines future states to provide dense, physically-grounded guidance for execution.

\noindent 3) We conduct extensive experiments demonstrating that Super-Mimic significantly outperforms state-of-the-art methods by over 20\% in complex, long-horizon tasks and validate the distinct advantages of our hierarchical video-based approach.

\section{RELATED WORK}
\subsection{Open-world Robotic Manipulation}
Open-world manipulation is crucial for achieving truly useful robots. Existing research can be broadly divided into two categories. The first category consists of methods based on Vision-Language-Action (VLA) models \cite{kim2024openvla, lin2025onetwovla, black2025pi0, zhou2025exploring, ma2024contrastive}. These methods typically rely on large-scale robotics data and extensive pre-training. Although they demonstrate strong generalization capabilities within known distributions, the high costs of data collection and training limit their adaptability in open-world environments.

The second line of work \cite{huang2023voxposer,liu2024moka,huang2024rekep, huang2022language, rana2023sayplan, song2023llm} explores zero-shot manipulation using MLLMs, requiring no additional training data. For instance, VoxPoser \cite{huang2023voxposer} and MOKA \cite{liu2024moka} leverage the semantic parsing and reasoning of Vision-Language Models (VLMs) to translate natural language instructions into executable conditional constraints or textual descriptions of keypoints. Similarly, ReKep \cite{huang2024rekep} uses a VLM to convert language descriptions into keypoint constraints. Beyond single-step or short-horizon manipulation, some works have also begun exploring long-horizon tasks. \cite{huang2022language, rana2023sayplan, song2023llm} attempt to complete more complex operational workflows through hierarchical planning or continuous reasoning. 
Complementary to these high-level planning frameworks, a significant body of research focuses on the robustness of physical interaction, particularly in robust and dexterous grasp detection. This includes developing methods for language-guided, task-oriented, and affordable 6-DoF grasping \cite{wei2025afforddexgrasp, wei2024grasp, wang2025task, wu2024economic, ma2024glover, DBLP:conf/icpr/JiangWCWZ24}, as well as extending manipulation to more complex scenarios like bimanual coordination \cite{DBLP:journals/corr/abs-2503-09186}.

Nevertheless, the aforementioned methods are primarily language-conditioned, meaning a robot's task execution is driven by language input. This approach can be challenging for complex, long-horizon tasks, as language is often too abstract to convey fine-grained operational constraints. To address this limitation, this research explores using human videos as a more intuitive and information-rich source of task information.

\begin{figure*}[htbp]
    \centering
    \includegraphics[width=1.0\linewidth]{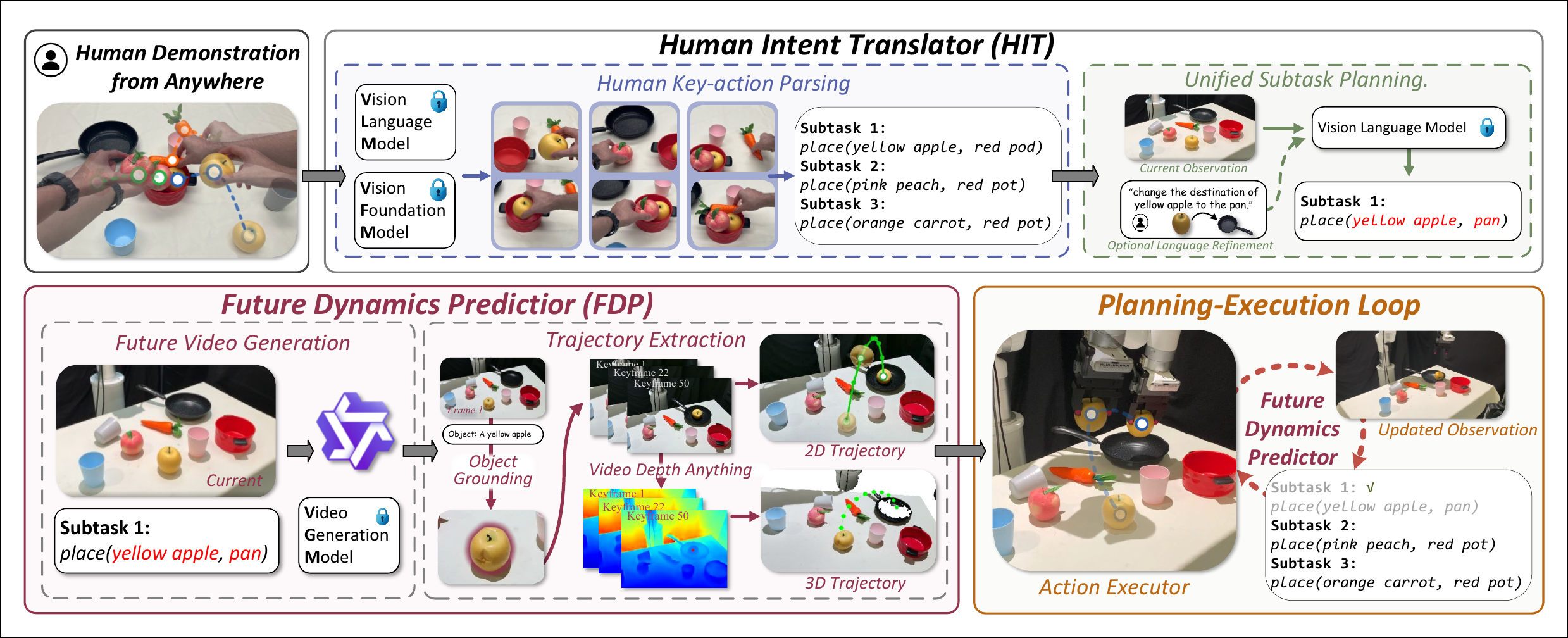}
    \caption{\textbf{Overview of Super-Mimic.} The HIT module uses a VLM to translate human demonstrations into an adaptable symbolic plan, enabling task modifications and skill transfers beyond simple imitation. Then the FDP module employs a video generation model to \textit{imagine} a plausible future execution for the current subtask. Finally, the Action Executor module grounds the imagined guidance into a final sequence of executable robot actions. Upon subtask completion, the system updates its observation and repeats the planning-execution loop.}
    \label{fig:overview}
\end{figure*}

\subsection{Learning from Human Demonstrations}
Learning long-horizon tasks from human videos is a promising yet challenging direction. Prior works have shown that demonstrations embedded in human videos contain rich semantic cues that can guide task execution \cite{mandlekar2020learning, wang2023mimicplay, guo2023learning, zhu2024vision, li2410okami}. Some approaches \cite{mandlekar2020learning, wang2023mimicplay} directly train models on large collections of human demonstrations to generate long-horizon action plans. However, this often requires extensive data collection and domain expertise, which is costly and limits scalability. 
Another line of research focuses on extracting short-horizon execution plans from a single human video. For example, \cite{guo2023learning} leverages point-cloud and pose estimation, ORION \cite{zhu2024vision} employs vision foundation models to extract manipulation actions and trajectories in a structured graph form, 
and OKAMI \cite{li2410okami} reformulates human actions for easier robot mapping. These methods then wrap the extracted trajectories into new environments with minor adjustments, generating execution plans for a new context.
Such methods offer an initial path toward adapting human demonstrations to robotic execution. Yet their reliance on low-level motion patterns challenges robustness across diverse environments, and directly mapping human videos onto the robot’s operating environment further limits their flexibility. This difficulty is partly due to the significant visual and morphological domain discrepancy between human demonstrators and robotic agents, a challenge that recent work has sought to mitigate through advanced visual pre-training strategies \cite{Zhou_2025_CVPR}.

In contrast, Super-Mimic leverages MLLMs to analyze human videos and perform multimodal subtask planning. 
This enables our system to go beyond simple imitation of human action sequences, allowing for both motion modifications and cross-environment transfers.
\subsection{Video Generation}
Video generation models utilize text or visual inputs to generate videos \cite{nvidia2025cosmos, wan2025, kong2024hunyuanvideo, polyak2024movie}, and a series of recent works have applied these models to the field of robot planning \cite{zhou2024robodreamer, fu2025learning, wu2023unleashing, bharadhwaj2024gen2act, ko2023learning,zhen2025tesseract}.
Some works focus on video generation models designed for robot manipulation. RoboDreamer \cite{zhou2024robodreamer} uses text-parsing to generate videos that capture complex spatial relationships and object interactions. RoboMaster \cite{fu2025learning} trains a robot manipulation video generation system by controlling the generation model with manipulation trajectories.
Other works concentrate on training modules to extract motion from pre-trained video generation models. GR-1 \cite{wu2023unleashing} takes language instructions, observed image sequences, and robot state sequences as multi-modal inputs to predict robot actions. Gen2Act \cite{bharadhwaj2024gen2act} trains a specialized trajectory translation model to bridge the gap between generated human videos and robot actions. AVDC \cite{ko2023learning} generates videos via a pre-trained optical flow network and inversely solves for robot actions, showing a degree of generalization in specific scenarios. Meanwhile, TesserAct \cite{zhen2025tesseract} fine-tunes a video generation model, proposing an effective world model framework that transforms RGB-D images of an entire scene into a spatio-temporally coherent 4D model.

Unlike methods that rely on separately trained trajectory translation modules, Super-Mimic operates in a fully zero-shot manner. Our FDP module leverages a general-purpose video depth estimator to directly extract 3D trajectories from its generated videos. This eliminates any need for task-specific data or training, significantly lowering the barrier to deployment in novel, unstructured environments.

\section{METHOD}

\subsection{Problem Formulation}

For long-horizon, zero-shot robotic manipulation in complex, open-world environments, robots must reason and act without task-specific training. This presents two fundamental challenges: high-level planning is often constrained by abstract language that fails to capture complex intent, while low-level control is limited by static observations that lack the dynamic information needed for physical execution. This work addresses these dual challenges by exploring richer visual modalities. We use unscripted human videos as a direct instructional signal for high-level planning, and prospectively ``imagined'' futures to provide a physically-grounded, dynamic reference for low-level action generation.

We consider the problem of long-horizon, open-world manipulation, denoted as task $\mathcal{T}$, where a robot must achieve a complex goal from high-level instructions in a zero-shot manner. The instructions, primarily composed of an unscripted human video demonstration $V$ and an optional natural language command $L$, are used to produce a sequence of subtasks, $\mathcal{P} = \{\tau_1, \dots, \tau_N\}$.
For each subtask $\tau_n \in \mathcal{P}$, we aim to generate a sub-plan $k_n$ from the robot's RGB-D observation $O_n$ in non-specific environments. The resulting sequence of sub-plans, $K = \{k_1, \dots, k_N\}$, provides the necessary spatio-temporal guidance for the action executor.

\subsection{System Overview}

To address the above problem, we propose \textbf{Super-Mimic}, a hierarchical framework that translates human demonstrations and text instructions into unified high-level plans, and utilizes video generation to \textit{imagine} future dynamics for reliable low-level execution, as illustrated in Figure \ref{fig:overview}.

Specifically, our framework is composed of two key modules that form a complete pipeline from abstract human intent to fine-grained robotic control. The first key module, the \textbf{Human Intent Translator}, leverages a VLM to translate rich human demonstrations into an adaptable symbolic plan. Beyond simple imitation of human actions, this enables task modifications and new skill transfers, making the framework applicable in various scenarios. The second key module, the \textbf{Future Dynamics Predictor}, bridges this high-level plan to the physical world by employing a video generation model to \textit{imagine} a plausible future execution for the current subtask, from which it distills a detailed, physically-grounded 3D trajectory. Finally, this trajectory is grounded by the Action Executor module, which translates the imagined guidance into a final sequence of executable robot actions. After a subtask is verified as complete, the system updates its observation and enters a new planning and execution loop.

\subsection{Human Intent Translator}

Given a human demonstration video $V$ depicting a long-horizon manipulation task, optionally with a language command $L$, the primary goal of high-level planning is to analyze the human's behavior and derive a corresponding task plan for the robot. However, raw, unscripted videos are complex and continuous, posing a significant challenge to extracting a deterministic, symbolic plan. To tackle this, the HIT module is designed to distill the core intent from the demonstration into a structured subtask plan $\mathcal{P}$.

\begin{figure}[htbp]
    \centering
    \includegraphics[width=1.0\linewidth]{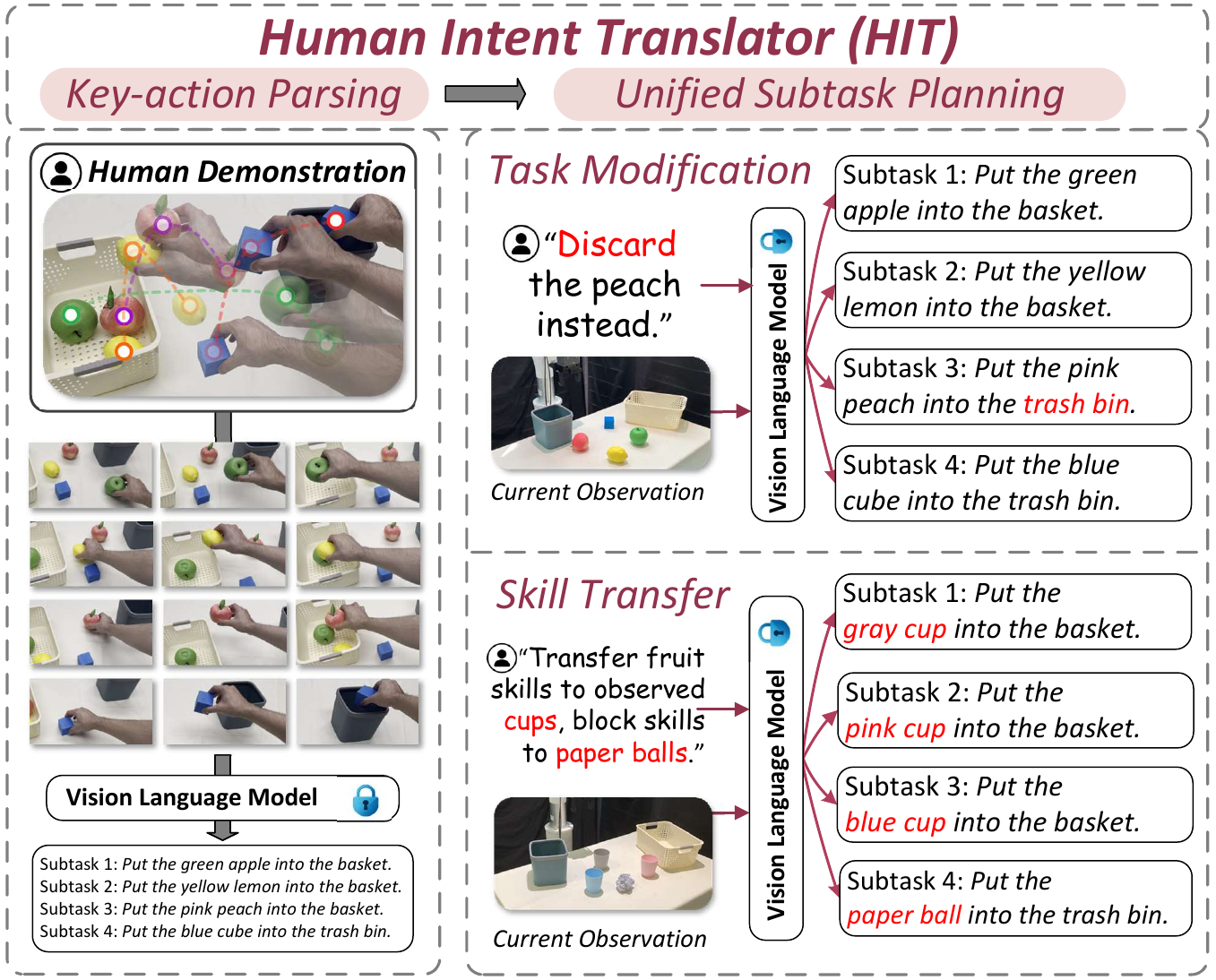} 
    \caption{The Human Intent Translator.}
    \label{fig:hit_modes}
\end{figure}

\noindent\textbf{- Human Key-action Parsing.}
The raw human video $V = \{f_1, \dots, f_T\}$, is distilled into a symbolic representation through two steps: keyframe extraction and demonstration abstraction.

First, the long video sequence is parsed into a sparse but semantically critical set of keyframes $V_{key}$. To achieve this, we apply a hand landmark detector (MediaPipe\cite{lugaresi2019mediapipe}) to each frame to track the 2D pixel coordinates of the wrist $p^{wrist}$. We identify keyframe candidates by detecting significant interaction events, which we assume are primarily indicated by stationarity. A frame $f_t$ is considered a candidate if the average L2 norm of the wrist's velocity within a local temporal window $W_t = \{t-\Delta t, \dots, t+\Delta t\}$ is below a threshold $\epsilon$:
$$
\frac{1}{|W_t|} \sum_{i \in W_t} ||p_i^{wrist} - p_{i-1}^{wrist}||_2 < \epsilon
$$
To prevent the selection of overly dense keyframes, candidates are temporally filtered to enforce a minimum frame interval, yielding the final keyframe sequence $V_{key}$.

Subsequently, these keyframes $V_{key}$ are passed to the VLM, which interprets the visual sequence and abstracts the demonstrated actions into a structured, descriptive \textit{baseline plan} $\mathcal{P}_{base}$. This abstraction is crucial as it transforms raw visual data into a compact symbolic representation. This reduces the computational load and focuses the subsequent planning stage on task-relevant actions.

\noindent\textbf{- Unified Subtask Planning.}
This stage takes the baseline plan $\mathcal{P}_{base}$, the robot’s current RGB observation $O^{rgb}$, and an optional language command $L$ as input to generate the final, executable plan $\mathcal{P}$. The system's behavior is conditioned by the language command $L$. If $L$ is absent, the system simply mimics the baseline plan by mapping the actions and objects in $\mathcal{P}_{base}$ to their analogues in the current scene. When a specific directive like $L=$ ``throw everything into the trash bin'' is provided, the VLM modifies the baseline plan to satisfy this language-guided constraint. This modification can range from a simple subtask edit to a more general skill transfer, where the system adapts learned manipulation knowledge to new objects.
The output of this entire process is the final plan $\mathcal{P} = \{\tau_1, \dots, \tau_N\}$, with each subtask $\tau_n$ represented as the tuple:
$$
\tau_n = (\text{desc}_n, \text{obj}_n, \text{loc}_n, \text{guide}_n, \text{precond}_n).
$$
where $\text{desc}_n$ is a natural language description of the subtask, $\text{obj}_n$ and $\text{loc}_n$ are semantic identifiers for the target object and its destination, $\text{guide}_n$ is the textual prompt used to guide the FDP's video generation, and $\text{precond}_n$ is a set of preconditions that must be met before execution.

In essence, the HIT module serves as the crucial bridge between ambiguous and complex human demonstrations and procedural task planning $\mathcal{P}$, enabling the robot to ground high-level instructions into an actionable plan.

\subsection{Future Dynamics Predictor}

The procedural plan $\mathcal{P}$ from the HIT module provides a logical roadmap of \textit{what} to do, but lacks the concrete, spatio-temporal guidance for \textit{how} to perform each subtask in the physical world. As illustrated in Figure~\ref{fig:overview}, the FDP module is designed to bridge this crucial ``symbol-to-physical" gap. Its core function is to take a symbolic subtask $\tau_n$ from the plan and enrich it with predictive dynamic information, to produce a detailed, physically-grounded keyframe trajectory $k_n$ that completes the subtask $\tau_n$.

\noindent\textbf{- Future Video Generation.}
For each subtask $\text{guide}_n$ in $\tau_n$, this stage leverages a video-generative model $G$ to forecast the task's execution. The model is conditioned on the robot's current observation $O_{n}^{rgb}$ and the textual prompt from $\text{guide}_n$. This process generates a short video sequence that imagines the physical outcome to complete the subtask $\tau_n$:
$$
V_n^{future} = G(O_{n}^{rgb}, \text{guide}_n)
$$
where $V_n^{future} = \{f'_1, f'_2, \dots, f'_Q\}$ is a sequence of imagined future frames. It provides a dynamic reference for subsequent motion extraction.

\noindent\textbf{- Trajectory Extraction.}
The imagined video $V_n^{future}$ is then analyzed through a multi-step pipeline to extract the structured keyframe trajectory $k_n$. First, an object tracker, Grounded SAM 2\cite{ren2024grounded}, identifies and tracks the target object through the video to produce a dense 2D pixel trajectory. The RDP algorithm\cite{douglas1973algorithms}  subsequently refines this path into a sparse set of 2D waypoints. To lift these waypoints into 3D, a video-based depth estimator, Video Depth Anything\cite{chen2025video}, generates temporally consistent depth maps from the video sequence. For each 2D waypoint, its depth is retrieved from the corresponding pre-computed map, and it is then projected to a 3D point $p^{3D}$ using the camera intrinsics.These signals are compiled into the trajectory $k_n = \{p_{1}^{3D}, \dots, p_{M}^{3D}\}$, a sequence of waypoints containing the spatio-temporal guidance for the Action Executor.

Ultimately, the FDP module enriches the procedural plan from HIT with predictive foresight, transforming an abstract subtask goal into a concrete, physically-grounded trajectory ready for execution.

\subsection{Action Executor}

The Action Executor is the final stage that grounds the imagined trajectory $k_n$ from the FDP into safe, physical robot actions. Originating from a holistically generated video, the trajectory $k_n$ is already spatially coherent and temporally aware of the scene's context. This provides a strong inductive bias for the subsequent motion planner, transforming the problem from one of path discovery to one of path refinement.

\noindent\textbf{- Grasp Planning and Trajectory Optimization.}
Before execution, the system performs two critical refinement steps. First, for grasp planning, AnyGrasp~\cite{fang2023anygrasp} is used to propose a set of stable candidates, and the MLLM selects the optimal one based on commonsense physical criteria. Second, the initial sequence of 3D waypoints $\{p_1^{3D}, \dots, p_M^{3D}\}$ from the FDP is refined by minimizing a weighted sum of a smoothness cost $C_{\text{smooth}}$ and a collision cost $C_{\text{coll}}$. The smoothness term, $C_{\text{smooth}} = \sum (1 - \cos\theta_m)$, penalizes the angle $\theta$ between consecutive path segments to prevent sharp turns. The collision term, $C_{\text{coll}} = \sum (\min_j \|p_m^{3D} - \mathbf{o}_j\|_2 + \varphi)^{-1}$, penalizes proximity to obstacle points $\mathbf{o}$ queried from a KD-Tree. This optimization is solved efficiently using a gradient-based local solver to find the refined trajectory.

\noindent\textbf{- Execution and Verification.}
The robot executes the optimized path. Afterward, a verification step uses a VLM model to assess the outcome against the subtask goal. After a subtask is verified as complete, the system updates its observation and enters a new planning and execution loop. Otherwise, control returns to the HIT module to initiate replanning based on the new observation.

\section{EXPERIMENTS}

The experiments are designed to rigorously evaluate the performance of our \textbf{Super-Mimic} framework on the long-horizon tasks and to dissect the contributions of its core modules. We aim to demonstrate that our ``Watch to Imagine" paradigm provides a more robust and flexible solution for long-horizon manipulation than existing text-driven approaches. To this end, our evaluation is structured to answer three central questions:

    \noindent \textit{Overall Performance:} How effectively does Super-Mimic solve complex zero-shot long-horizon tasks using human demonstrations, compared to text-only baselines?
    
    \noindent \textit{The contribution of ``Watching":} What is the quantifiable advantage of parsing human demonstrations with our HIT module compared to relying solely on language instructions?
    
    \noindent \textit{The significance of ``Imagining":} What is the specific contribution of our FDP module to the system's execution robustness and overall success?

\subsection{Experimental Setup}
\begin{figure}[htbp]
    \centering
    \includegraphics[width=0.9\linewidth, trim={0cm 6cm 0cm 0cm}, clip]{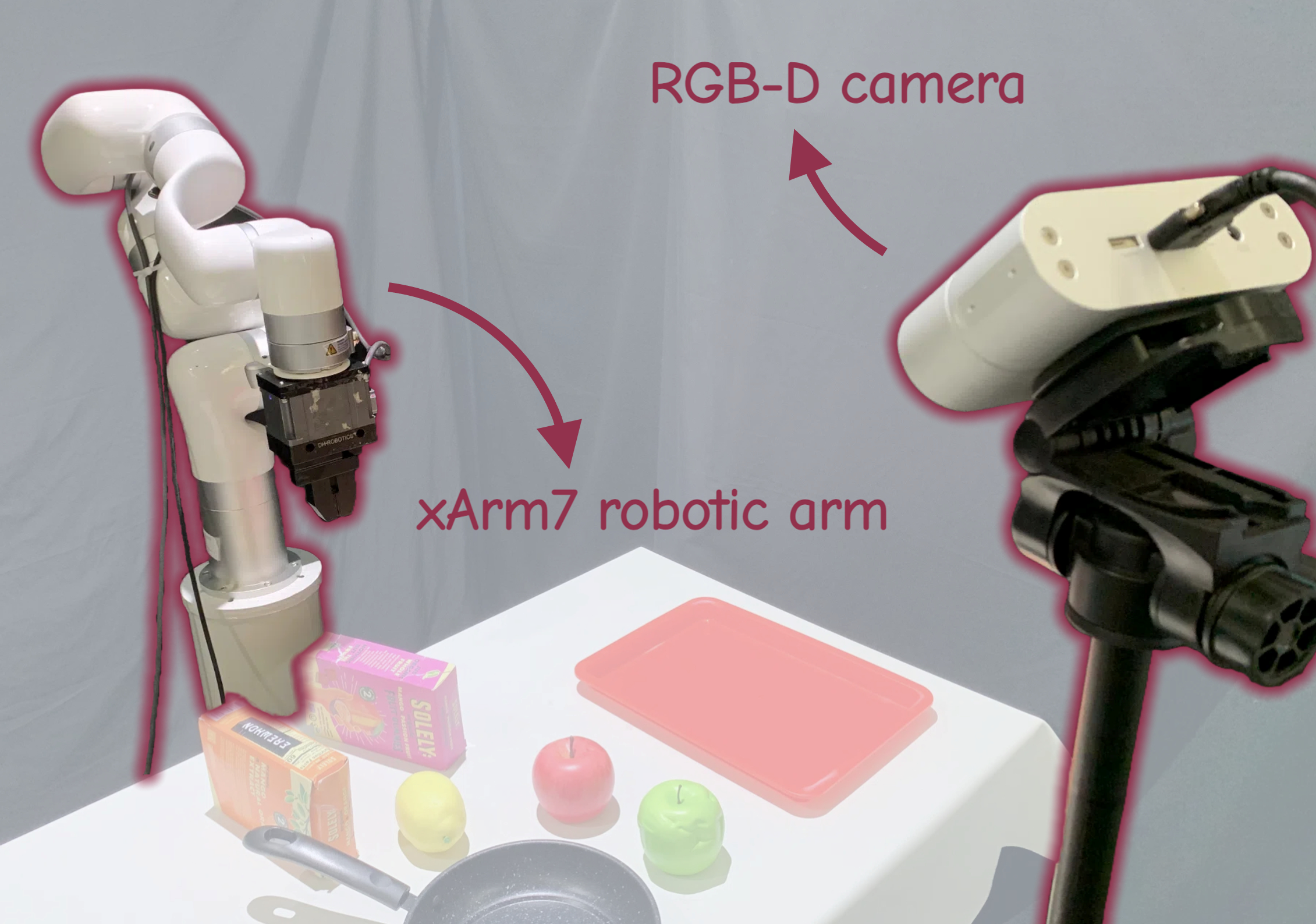}
    \caption{Our experimental platform, consisting of a 7-DoF xArm7 arm and a third-view Orbbec RGB-D camera.}
    \label{fig:setup}
\end{figure}

Our real-world experiments are conducted on a manipulation platform featuring a 7-DoF xArm7 robotic arm and an externally mounted Orbbec Femto Bolt camera for RGB-D perception, as depicted in Fig.~\ref{fig:setup}. Our framework is implemented using several off-the-shelf models for its core perceptual and generative functions: Wan2.2-Lightning~\cite{wan2025} serves as the model for future video generation, MediaPipe\cite{lugaresi2019mediapipe} is used for hand landmark detection, and Grounded SAM 2~\cite{ren2024grounded} is employed for object tracking. To ensure a fair and consistent comparison, all high-level planning and reasoning tasks across our entire evaluation, for both Super-Mimic and the baselines, are powered by the Qwen2.5-VL 72B~\cite{Qwen2.5-VL} VLM.

\subsection{Experimental Settings} To thoroughly evaluate our method, we test Super-Mimic under four distinct instruction settings, which are also presented in Table \ref{table:main_results}:

    \noindent \textit{(1) Short Simple language}: The system operates without video, relying solely on a brief text instruction that specifies the entire long-horizon task.

    \noindent \textit{(2) Video}: The system uses only a human video to infer and execute the task in an analogous environment.
    
    \noindent \textit{(3) Video + Constraints}: This setting combines a human demonstration video with supplementary language constraints to modify the demonstrated task.
    
    \noindent \textit{(4) Skill Generalization}: The most challenging setting, where the system must abstract the core skill from a human demonstration (e.g., sorting) and apply it to a novel scene with completely different objects.

\subsection{Tasks, Metrics, and Baselines}

\noindent\textbf{- Tasks:} To test distinct capabilities, we evaluate our framework across three challenging long-horizon scenarios, each comprising 5 to 8 subtasks that demand high execution fidelity to prevent cascading failures:

    \noindent \textit{Meal Preparation}: A 5-subtask sorting task requiring the robot to categorize fruits and snack boxes into different containers.
    
    \noindent \textit{Tidy Up the Desk}: A 5-subtask task where the robot must use commonsense reasoning to sort trash from useful items into a bin and a basket.
    
    \noindent \textit{Irregular Traversal}: An 8-subtask task with non-logical pick-and-place motions, designed to be difficult to describe with language and to challenge goal-oriented planners.

\noindent\textbf{- Metrics:} We evaluate over $N=20$ trials using two metrics:
\textbf{Task Success Rate (TSR)} and \textbf{Subtask Success Rate (SSR)}, defined as:
\begin{align*}
\mathrm{TSR} &= \frac{1}{N} \sum_{i=1}^{N} S_i, \\
\mathrm{SSR} &= \frac{1}{N} \sum_{i=1}^{N} \left( \frac{n_i}{M} \right)
\end{align*}
where for each trial $i$:  
$S_i = 1$ if all subtasks succeeded, else $0$;  
$n_i$ is the number of successfully completed subtasks;  
$M$ is the total number of subtasks in the task.

\noindent\textbf{- Baselines:} We compare against two state-of-the-art text-driven methods, ReKep~\cite{huang2024rekep} and MOKA~\cite{liu2024moka}. For a fair comparison, we use their official implementations, adapting only the perception and execution modules to our hardware while retaining their core reasoning and planning pipelines.
ReKep~\cite{huang2024rekep} is a framework that prompts a VLM to generate robot behaviors by producing optimizable, spatio-temporal constraints as Python functions; a numerical solver then generates actions by satisfying these constraints. MOKA\cite{liu2024moka} employs a mark-based visual prompting technique, converting the motion generation problem into a visual question-answering task where a VLM selects affordance keypoints and waypoints from a 2D image.

Since both methods are designed for text-only instruction and do not support video input, we evaluate them under two distinct instruction protocols:

    \noindent\textit{(1) Short Simple Language}: The baseline receives a single, high-level natural language instruction for the entire task, following their original practice.
    
    \noindent\textit{(2) Long Detailed Language}: The task is manually pre-decomposed into a sequence of simpler commands, which are provided to the baselines one by one. While our method parses instructions directly from complex human video demonstrations, the baselines receive textual inputs that are highly detailed and provide precise guidance to ensure a fair comparison.

\subsection{Comparative Analysis}

\begin{table*}[htbp]
\caption{Overall performance comparison on three tasks. (TSR / SSR \%)}
\label{table:main_results}
\centering
\small
\setlength{\tabcolsep}{6pt}

\begin{tabular}{l|l|c|c|c}
\toprule
\textbf{Method} & \textbf{Setting} & \textbf{Meal Preparation} & \textbf{Tidy Up} & \textbf{Irregular Traversal} \\
\midrule
\multirow{2}{*}{\textbf{ReKep}~\cite{huang2024rekep}} 
  & Short Simple Language & 0 / 14 & 0 / 8 & N/A \\
  & Long Detailed Language & 10 / 44 & 10 / 50 & 0 / 20 \\
\midrule
\multirow{2}{*}{\textbf{MOKA}~\cite{liu2024moka}} 
  & Short Simple Language & 0 / 26 & 10 / 30 & N/A \\
  & Long Detailed Language & 20 / 60 & 20 / 64 & 0 / 26 \\
\midrule
\multirow{4}{*}{\textbf{Super-Mimic (Ours)}} 
  & Short Simple Language & \textbf{30 / 76} & \textbf{30 / 74} & N/A \\
  & Video & \textbf{50 / 82} & \textbf{40 / 78} & \textbf{20 / 58} \\
  & Video + Constraints & \textbf{40 / 84} & \textbf{30 / 64} & \textbf{20 / 42} \\
  & Skill Generalization & \textbf{30 / 68} & \textbf{30 / 58} & N/A \\
\bottomrule
\end{tabular}
\end{table*}

The results of our comparative evaluation are summarized in Table \ref{table:main_results}, which reveals key findings regarding the superiority of video demonstrations, the challenges of text-only instruction on long-horizon planning, and the unique generalization capabilities of our framework.

\noindent\textbf{- Text-Only Long-Horizon Planning}
In the text-only setting, the results reveal a primary bottleneck for baseline methods: autonomous task decomposition. Both ReKep~\cite{huang2024rekep} and MOKA\cite{liu2024moka} show a considerable performance increase when provided with detailed, step-by-step language compared to a single, simple command (e.g., MOKA's TSR on Tidy Up the Desk rises from 10\% to 20\%). This confirms their difficulty in independently planning a multi-step procedure from an abstract goal. Notably, Super-Mimic, when given the same challenging simple command, achieves a 30\% TSR on the same task—a result that not only surpasses the baselines' performance with simple commands but is also superior to their results even when they were aided by detailed manual instructions.

\noindent\textbf{- Superiority of Video Demonstrations}
Super-Mimic's video-guided approach significantly outperforms the baselines' best-case scenarios. In the Video setting, our method achieves a 50\% TSR in Meal Preparation, more than doubling the 20\% TSR of the strongest baseline (MOKA\cite{liu2024moka} with detailed instructions). This performance gap highlights a fundamental limitation of text-driven methods: they must infer complex spatial goals from sparse language, a process prone to ambiguity. A video demonstration, in contrast, provides a direct, rich visual signal that allows our HIT module to ground the task's intent and spatial nuances with much higher fidelity.

\noindent\textbf{- Flexibility and Generalization ability}
Our framework demonstrates unique flexibility and generalization capabilities unlocked by video. The Video + Constraints setting maintains a high success rate (e.g., 40\% TSR on Meal Preparation), showing that the system can flexibly refine demonstrated skills with new text commands. Most notably, the success in the Skill Generalization setting, with a 30\% TSR on complex sorting tasks, validates that our framework can abstract the underlying logic of a task and reapply it in novel contexts—a crucial step towards general-purpose manipulation.

\noindent\textbf{- Robustness in Ambiguous Scenarios.}
The "Irregular Traversal" task was specifically designed to challenge the limits of language-based planning. It involves a sequence of actions with no clear semantic logic, making it exceptionally difficult to describe with text. As a result, both ReKep~\cite{huang2024rekep} and MOKA\cite{liu2024moka} fail entirely on this task, even with detailed manual instructions. In contrast, Super-Mimic, by mimicking the visual demonstration and then imagining the trajectory, achieves a 20\% TSR. This result showcases the unique robustness of our video-driven paradigm for tasks that defy simple linguistic or geometric description.

\subsection{Ablation Studies}
To dissect the contributions of our core modules, we conduct a series of ablation studies to isolate the impact of "Watching" (HIT) and "Imagining" (FDP) on overall performance. The results are summarized in Table \ref{table:ablation_results}.

\begin{table}[htbp]
\caption{Ablation study results (TSR / SSR \%)}
\label{table:ablation_results}
\centering
\begin{tabular}{lccc}
\toprule
\textbf{Setting} & \textbf{Meal Prep.} & \textbf{Tidy Up} & \textbf{Irregular Trav.} \\
\midrule
\textbf{Super-Mimic (Full)} & \textbf{50 / 82} & \textbf{40 / 78} & \textbf{20 / 58} \\
(1) w/o FDP & 20 / 58 & 20 / 62 & 0 / 24 \\
(2) w/o Path Guidance & 30 / 72 & 30 / 74 & 20 / 56 \\
(3) w/o HIT (Short Lang.) & 30 / 76 & 30 / 74 & 0 / 32 \\
(4) w/o HIT's Parsing & 40 / 70 & 30 / 70 & 10 / 44 \\
\bottomrule
\end{tabular}
\end{table}

\noindent\textbf{- Analysis of the Future Dynamics Predictor}
The contribution of "Imagining" is evaluated in ablations (1) and (2). In (1) (\textit{w/o FDP}), we replace the FDP with a static planner that selects 2D waypoints from the current observation, following the methodology of MOKA~\cite{liu2024moka}. This leads to a drastic performance decline (e.g., TSR for Meal Preparation drops from 50\% to 20\%), confirming that static planning is insufficient. To further dissect FDP's contribution, in (2) (\textit{w/o Path Guidance}), we provide the planner with only the final destination point from the imagined video. While this improves performance over the fully static planner, it still underperforms our full pipeline. These results jointly demonstrate that \textit{imagining} the future is critical, and that the \textit{complete trajectory} provided by FDP, not just the endpoint, is key to robust execution.

\noindent\textbf{- Analysis of the Human Intent Translator}
The value of "Watching" is analyzed in ablations (3) and (4). Ablation (3) (\textit{w/o HIT}) measures the performance in a text-only setting, directly quantifying the significant performance gain from using video demonstrations (e.g., TSR on Tidy Up jumps from 30\% to 40\%). Ablation (4) (\textit{w/o HIT's Parsing Step}) investigates the importance of our parsing module by feeding the full, raw video to the VLM planner. The resulting performance degradation (e.g., TSR on Irregular Traversal drops from 20\% to 10\%) proves that HIT's key-action parsing is not just a trivial pre-processing step, but a crucial component for distilling clear, actionable intent from noisy and redundant video signals.

\subsection{Failure Analysis}
\begin{figure}[thpb]
    \centering
    \parbox[t]{0.32\linewidth}{\centering
         \includegraphics[width=0.9\linewidth]{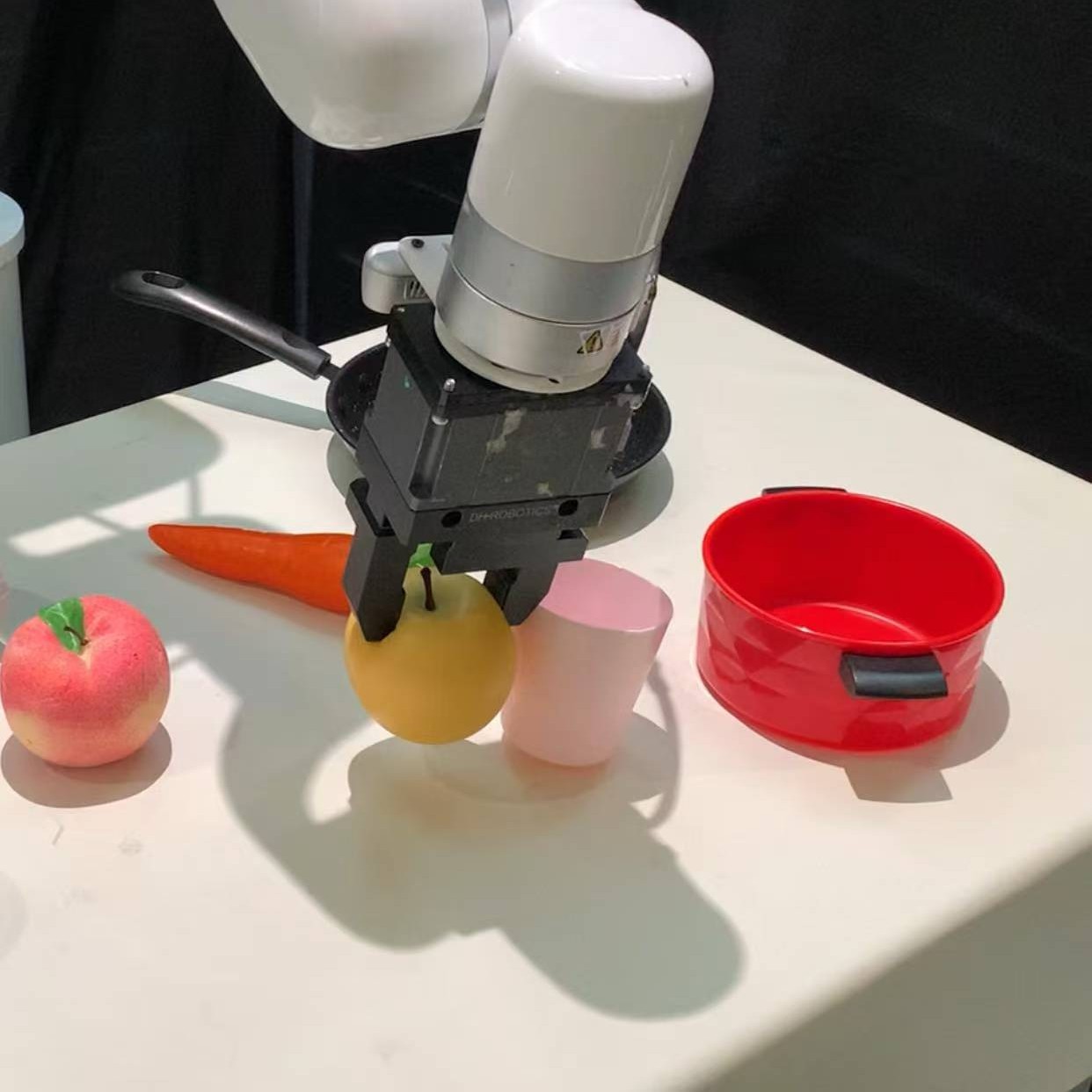}
        \centerline{(a) HIT Failure}}
    \hfill
    \parbox[t]{0.32\linewidth}{\centering
        \includegraphics[width=0.9\linewidth]{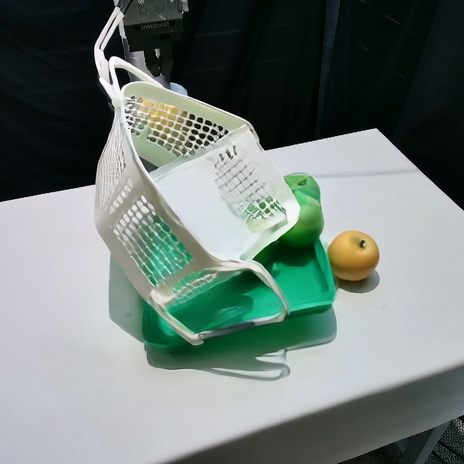} 
        \centerline{(b) FDP Failure}}
    \hfill
    \parbox[t]{0.32\linewidth}{\centering
        \includegraphics[width=0.9\linewidth]{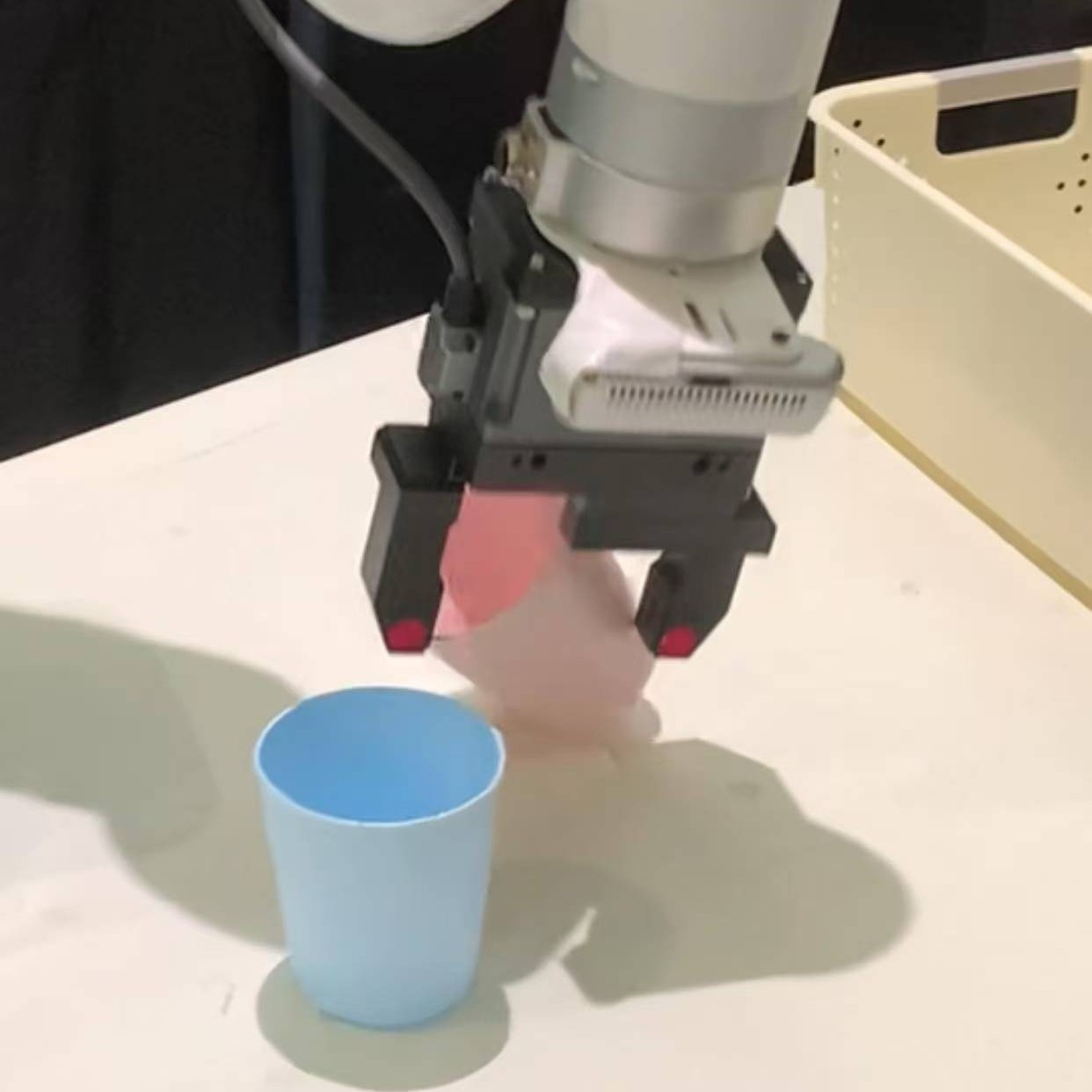}
        \centerline{(c) Execution Failure}}
    \caption{Examples of major failures: (a) HIT planning error (e.g., misplacing the apple), (b) FDP prediction error (e.g., implausible object deformation), (c) Execution error (most common, e.g., grasp failure knocks over the cup).}
    \label{fig:failure_examples}
\end{figure}

To provide a nuanced understanding of our system's limitations, we analyze its three primary failure modes (Fig. \ref{fig:failure_examples}). The first, HIT Planning Failure, occurs when the VLM misinterprets user intent, for example, incorrectly planning to place an apple into a cup (Fig. \ref{fig:failure_examples}a). The second, FDP Prediction, involves the video model hallucinating a physically implausible future, such as a completely deformed object (Fig. \ref{fig:failure_examples}b), resulting in an unusable trajectory. The final category is Execution Failure, where the low-level physical interaction fails despite a correct plan. This is often caused by perception inaccuracies or imprecise grasping, leading to outcomes like knocking over an object (Fig. \ref{fig:failure_examples}c).

\section{CONCLUSIONS and DISCUSSIONS}
\noindent\textbf{Conclusion.}
In this paper, we presented \textbf{Super-Mimic}, a hierarchical framework for zero-shot, long-horizon robotic manipulation. We establish a ``Watch to Imagine" paradigm where the HIT module parses human videos into structured plans, and the FDP module leverages video generation to imagine subtask execution, providing physically-grounded guidance for control. Experiments demonstrate the effectiveness of our approach on complex, long-horizon tasks. We believe this paradigm presents a promising direction for building general-purpose robots that learn complex skills from observing humans.

\noindent\textbf{Discussion.}
Despite promising results, our work with Super-Mimic points to several avenues for future research, reflecting common challenges across the field. Specifically, the fidelity of our imagined trajectories hinges on the capabilities of current video generation models. While these models are powerful for visual synthesis, they are not yet optimized for the deterministic, high-precision predictions essential for robotics. Furthermore, as with any physically embodied system, our framework's performance is constrained by the accuracy of its perception. Finally, advancing beyond sequential task plans to more flexible, non-linear replanning frameworks also remains a key open problem for long-horizon robotics. We are optimistic that continued progress in these underlying technologies will further enhance the capabilities of Super-Mimic and other similar frameworks.



\bibliographystyle{IEEEtran}
\bibliography{reference}

\end{document}